\documentclass[letterpaper, 10 pt, conference, dvipsnames]{ieeeconf}  

\IEEEoverridecommandlockouts                              
\usepackage{graphicx}
\usepackage{tabularx}
\usepackage[utf8]{inputenc}
\usepackage[T1]{fontenc}
\usepackage{subcaption}

\newsavebox{\twosubbox}
\usepackage{xcolor, hyperref}
\usepackage{color,soul}
\usepackage{listings}
\usepackage{xcolor}
\usepackage{placeins}
\usepackage{float}

\definecolor{codegreen}{rgb}{0,0.6,0}
\definecolor{codegray}{rgb}{0.5,0.5,0.5}
\definecolor{codepurple}{rgb}{0.58,0,0.82}
\definecolor{backcolour}{rgb}{0.95,0.95,0.92}
\definecolor{hyellow}{rgb}{0.97,0.98,0.71}
\definecolor{hred}{rgb}{0.97,0.82,0.76}
\definecolor{hblue}{rgb}{0.76,0.93,0.97}

\lstdefinestyle{codestyle}{
    backgroundcolor=\color{backcolour},   
    commentstyle=\color{codegreen},
    keywordstyle=\color{magenta},
    numberstyle=\tiny\color{codegray},
    stringstyle=\color{codepurple},
    basicstyle=\ttfamily\footnotesize,
    breakatwhitespace=false,         
    breaklines=true,                 
    captionpos=b,                    
    keepspaces=true,                 
    numbersep=5pt,                  
    showspaces=false,                
    showstringspaces=false,
    showtabs=false,                  
    tabsize=2,
    escapechar=@
}

\lstdefinelanguage{json}{
    basicstyle=\normalfont\ttfamily,
    numbers=left,
    numberstyle=\scriptsize,
    stepnumber=1,
    numbersep=8pt,
    showstringspaces=false,
    breaklines=true,
    frame=lines,
    backgroundcolor=\color{background},
    literate=
     *{0}{{{\color{numb}0}}}{1}
      {1}{{{\color{numb}1}}}{1}
      {2}{{{\color{numb}2}}}{1}
      {3}{{{\color{numb}3}}}{1}
      {4}{{{\color{numb}4}}}{1}
      {5}{{{\color{numb}5}}}{1}
      {6}{{{\color{numb}6}}}{1}
      {7}{{{\color{numb}7}}}{1}
      {8}{{{\color{numb}8}}}{1}
      {9}{{{\color{numb}9}}}{1}
      {:}{{{\color{punct}{:}}}}{1}
      {,}{{{\color{punct}{,}}}}{1}
      {\{}{{{\color{delim}{\{}}}}{1}
      {\}}{{{\color{delim}{\}}}}}{1}
      {[}{{{\color{delim}{[}}}}{1}
      {]}{{{\color{delim}{]}}}}{1},
}

\newcommand{\invisible}[1]{}

\title{\LARGE \bf
LMPVC and Policy Bank: Adaptive voice control for industrial robots with code generating LLMs and reusable Pythonic policies
}

\author{Ossi Parikka$^{1}$ and Roel Pieters$^{1}$
\thanks{$^{1}$Cognitive Robotics group, Unit of Automation Technology and Mechanical Engineering, Tampere University, 33720, Tampere, Finland; {\tt\small firstname.surname@tuni.fi}}%
}

\begin{document}

\maketitle
\thispagestyle{empty}
\pagestyle{empty}

\begin{abstract}


Modern industry is increasingly moving away from mass manufacturing, towards more specialized and personalized products. As manufacturing tasks become more complex, full automation is not always an option, human involvement may be required. This has increased the need for advanced human robot collaboration (HRC), and with it, improved methods for interaction, such as voice control. Recent advances in natural language processing, driven by artificial intelligence (AI), have the potential to answer this demand. Large language models (LLMs) have rapidly developed very impressive general reasoning capabilities, and many methods of applying this to robotics have been proposed, including through the use of code generation. This paper presents \textit{Language Model Program Voice Control} (LMPVC), an LLM-based prototype voice control architecture with integrated policy programming and teaching capabilities, built for use with Robot Operating System 2 (ROS2) compatible robots. The architecture builds on prior works using code generation for voice control by implementing an additional programming and teaching system, the \textit{Policy Bank}. We find this system can compensate for the limitations of the underlying LLM, and allow LMPVC to adapt to different downstream tasks without a slow and costly training process. The architecture and additional results are released on GitHub (\url{https://github.com/ozzyuni/LMPVC}).

\end{abstract}

\section{Introduction}\label{sec:intro}

\begin{figure*}[tp]
    \centering
    \includegraphics[width=\textwidth]{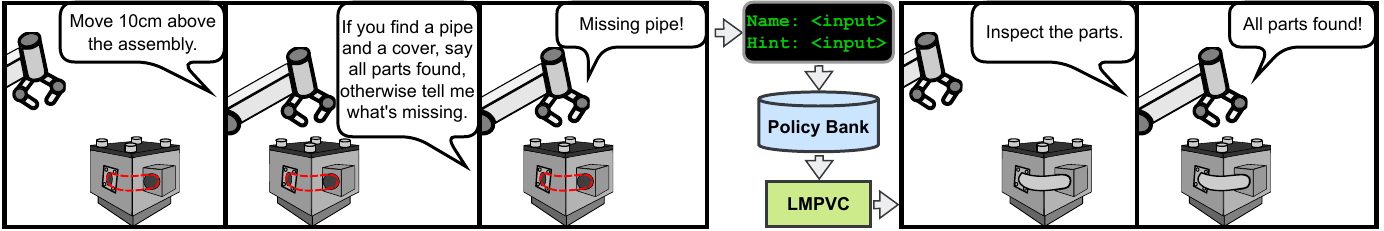}
    \caption{With the Policy Bank, a user can instruct the robot to complete a task step by step, and save the resulting action code as a new Policy. This Policy is then integrated into the prompting structure of LMPVC, allowing it to be intelligently recalled when needed. Policies can also be programmed, edited and combined, making the system arbitrarily expandable.}
    \label{fig:title}
\end{figure*}

In the EU and other similar areas, where mass manufacturing is no longer viable in the global market due to factors such as labor costs, industry is looking towards increasingly complex and customizable products as a way to the future. To make this future possible, industrial robots face new, challenging requirements. In particular, these developments have exposed the need for human-robot collaboration (HRC) on a level beyond what current technology allows. \cite{inkulu2022caoi,kumar2021hrcsurvey,bauer2016lightweight} Improvements are needed in many areas, one of which is more natural interaction through voice control \cite{Marge_2022}.

Natural language understanding in the context of voice controlled robots may be considered as a problem of \textit{grounding} human-intelligible linguistic concepts in the space of robot actions and perception. Reducing the entirety of human expression into a finite set of actions is quite a challenging problem, but recent advances in language modeling using artificial intelligence (AI) present new avenues for solving it \cite{fang2025sam2act,goyal2024rvt2,driess2023palme,singh2023progprompt}. One such system is \textbf{Code as Policies} (CaP) \cite{liang2023code}, which introduced a method for grounding natural language with code generation, using pre-trained Large Language Models (LLMs). The CaP model performs well for many general tabletop manipulation tasks by interacting with an open vocabulary object detector, but this combination alone is not directly applicable to an industrial setting. Open vocabulary detectors may be fine-tuned to a specific industrial scope if necessary, although adopters might prefer to keep using whatever system they might already have. More importantly, many industrial tasks involve more than manipulation, and an LLM often lacks the necessary knowledge to effectively answer prompts such as "Inspect the part" or "Weld the bottom seam".

Some fragments of a solution are already present in CaP, in the form of the possibility to make use of libraries and, under limited circumstances, previously generated code. Vemprala et. al \cite{vemprala2023chatgptroboticsdesignprinciples} further hint at the possibility of using the latter to teach the robot progressively more complex tasks. Based on a previous M.Sc. thesis \cite{lmpvcThesis}, this paper seeks to exploit these possibilites to improve the interactivity and adaptability of industrial voice control systems, as well as to provide tools for easier prototyping with real robots and local LLMs. To this end, we propose the following:

\begin{enumerate}
    \item \textbf{LMPVC:} Open, modular, LLM-based voice control architecture with support for ROS 2.
    \item \textbf{Policy Bank:} Framework for creating Pythonic policies through voice-based teaching and programming, and automatically prompting an LLM to use them. 
\end{enumerate}

These systems create a simple and effective approach to adapt voice control to new domains through learning from descriptive commands, as summarized in Fig. \ref{fig:title}. 

\section{Related Work}\label{sec:related_work}

The rapid advancement of artificial intelligence in recent years has opened the floodgates for research on new approaches to natural language human-robot interaction (HRI). The catalyst for this revolution, \textbf{Large Language Models}, introduced the world to machines with remarkable capabilities in natural language understanding and language-based general reasoning \cite{teubner2023welcomechatgpt}. However, as the name suggest, LLMs such as ChatGPT predominantly work in the realm of language, with additional work required to bridge their capabilities with other domains.

Initially, the application of LLMs to robotics reflected this divide, with the most natural direction being in the realm of social robotics, where LLMs could provide social interaction with text and speech. More recently, advances in models have made them more desirable for instruction following and task completion purposes as well \cite{ZhangLlmHriReview}. The approaches to bridging the language-action cap may be seen as two-fold: On the one hand, dedicated models using technology from LLMs have been developed to work in the robotics domain directly, and on the other, various methods have been discovered to apply LLMs as they are.

In the first case, solutions commonly involve \textbf{Vision-Language Models} (VLM) and \textbf{Large Vision Models} (LVM), which deploy Transformer architectures similar to LLMs, but adjusted to also process visual data. An earlier solution is CLIPort \cite{shridhar2021cliport}, which combines existing visual and robotics models for an end-to-end solution. By contrast, recent solutions such as RVT-2 \cite{goyal2024rvt2}, SAM2Act \cite{fang2025sam2act} and others \cite{duan2024manipulateanythingautomatingrealworldrobots,shridhar2022perceiveractormultitasktransformerrobotic} are fully developed with robotics in mind, and further improve visual precision by employing methods such as using point clouds to render the scene from different angles instead of relying on cameras alone. SAM2Act is currently state of the art in RLBench with an average score of 86.8\%, showcasing impressive performance in complex manipulation tasks, but despite these results, it and other similar models are not without disadvantage. Notably, because the models are end-to-end and entirely bespoke, it can be difficult to interact with how they operate, or add functionality which they do not possess. RVT-2 and SAM2Act do both involve few-shot learning capabilities, but this is not an ideal solution to all problems.

On the side of using LLMs more directly, much progress has also been made. Text2Motion \cite{lin2023t2m} uses PDDL to allow an LLM to reason about an environment in terms of symbolic states and known skills. RobotGPT \cite{jin2023robotgpt} uses ChatGPT to distill a smaller, dedicated model also based on generating executable actions, achieving higher stability but sacrificing some flexibility. \textbf{Code as Policies} \cite{liang2023code} uses OpenAI Codex to generate robot code from natural language prompts, \textit{grounding} natural language to a set of executable queries and actions. CaP can also leverage its code-based nature by incorporating libraries and hierarchically building behaviour from individual functions. ProgPrompt \cite{singh2023progprompt} similarly takes advantage of code generation for task planning. Vemparala \textit{et al.} \cite{vemprala2023chatgptroboticsdesignprinciples} further investigate the collaborative aspects of using ChatGPT to generate code, allowing the user to request changes to generated tasks. Some, such as PaLM-E \cite{driess2023palme} also incorporate visual data more directly through encoding. In comparison to VLMs, these LLM based solutions place less emphasis on environmental modeling and precise manipulation, but more on interactivity and the role of the human operator. They also offer more points of entry for programmers, and do not necessarily require the training of task-specific models.

\section{Problem Definition}\label{sec:problem}

Recent dedicated robotics models such as SAM2Act are highly effective at precisely grounding natural language in various manipulation tasks, but may not be as easily applicable to specialized industrial use cases. It is not trivial to precisely predict or modify their behaviour, and their bespoke nature requires potential users to commit to a specific system that might not meet all future needs. For industrial actors looking for stability and configurability, this can be a difficult proposition. For the moment, the flexibility and upgradibility advantages of an LLM based solution may be worth the compromises, provided they can be adapted to perfom the task actually required by a given industrial application. 

Fortunately, the code-based approach of Code as Policies offers many opportunities for implementing adaptable and specialized behaviour. Instead of purely relying on model performance, we believe a hybrid approach with first party libraries and reused generated code may hold advantages not fully explored by CaP. Furthermore, most existing solutions, including CaP, target SotA cloud-based models, with little regard for latency or data security. In real industrial applications these can be critical, and to better answer this need, the focus of this paper will instead be on LLMs available for local deployment. Finally, we find that such research could benefit from more freely available tooling for rapid prototyping. \cite{liang2023code}

\section{System}\label{sec:design}

\begin{figure*}[tp]
    \centering
    \includegraphics[width=\textwidth]{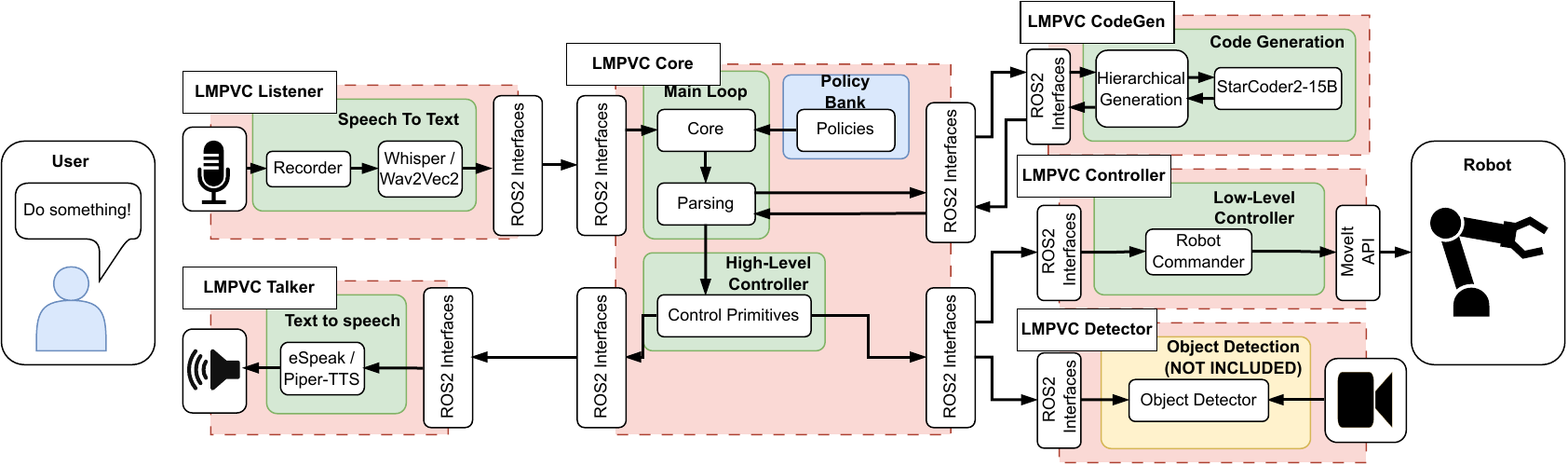}
    \caption{The modular structure of the LMPVC architecture encourages modification and increases adaptability.}
    \label{fig:lmpvc}
\end{figure*}

The LMPVC architecture is designed to be modular, with individual components being implemented separately using their own interfaces, as shown in Fig. \ref{fig:lmpvc}. Most components also reside in their own Robot Operating System (ROS) 2 packages and nodes. This structure allows various components to be modified or rebuilt entirely without affecting the rest of the system, provided the existing interfaces of the modules are respected. Some consideration has also been given to uses outside of ROS: Where possible, ROS-specific code is isolated from internal logic, and if desired, can in many cases be replaced entirely by direct calls between Python modules. The object detection module is a special case, as the reference implementation only contains interfaces, which can be adapted to any common object detection model as necessary. 

\subsection{Control flow}

At the center of the architecture is the Main Loop, contained within the Core module. In typical usage, the Core first triggers the Listener after detecting a wake event, such as a key press. It waits until the Listener returns transcribed text, or a timeout in case of no detected speech. LMPVC provides two engines by default, Wav2Vec 2.0 \cite{baevski2020wav2vec20frameworkselfsupervised} and Whisper \cite{radford2022whisper}, of which the latter is generally recommended. Any returned text is filtered for manually detected keywords which initiate actions directly, such as "stop", and if none are found, passed to Code Generation. Commands for controlling the teaching capabilities of the Policy Bank are also currently implemented as keywords, although there is no technical reason this could not be done from generated code through dedicated primitives. 

Once the command is parsed and necessary code has been generated, the High-Level Controller component of the Core module takes over for execution of the high-level control primitives. Different primitives can involve moving the robot through the Controller module, recovering object poses from the Detector module, or providing user feedback with the Talker module. Optionally, generated code can be temporarily retained as context for future actions.

\subsection{Code generation}

Code Generation is based on Code as Policies: As illustrated in Fig. \ref{fig:cap}, the model is prompted with a static \colorbox{hred!50}{preamble} containing examples, combined with a user \colorbox{hyellow!70}{prompt} formatted as a comment. The LLM then generates an executable \colorbox{hblue!50}{response}. Importantly, our solution replaces OpenAI Codex with a locally running LLM, StarCoder2 \cite{lozhkov2024starcoder2stackv2}. Liang \textit{et al.} refer to the results of complete generation cycles as Language Model Programs (LMP), a practice we adopt to distinguish between LMPs and the policies handled by the Policy Bank. \cite{liang2023code}

\begin{figure}[h!]

\begin{lstlisting}[language=Python, backgroundcolor=\color{hred!50}, belowskip=0pt]
#define function: say thank you 
def say_thank_you(robot):
    robot.say("thank you")
\end{lstlisting}
\begin{lstlisting}[language=Python, backgroundcolor=\color{hyellow!70}, aboveskip=3pt, belowskip=3pt]
#define function: tell me the first law of robotics
\end{lstlisting}
\begin{lstlisting}[language=Python, backgroundcolor=\color{hblue!50}, aboveskip=0pt]
def tell_me_the_first_law_of_robotics(robot):
    robot.say("A robot may not injure a human being, or through inaction, allow a human being to come to harm.")
#end of function
\end{lstlisting}

\caption{Basic operating principle of the code generation.}
\label{fig:cap}
\end{figure}

Beyond simple control primitives such as \textit{say} or \textit{move}, CaP demonstrates that generated code can depend on both first- and third-party libraries for more complex functionality, such as using NumPy for mathemathical operations. Perhaps most interestingly, because the model works by generating code, it can also make use of its own outputs under certain circumstances. In CaP, this is used in two ways: hierarchical generation and LMP composition. We implement hierarchical generation to improve the performance of code generation, particularly in complex tasks. This works by prompting the model to call \textit{undefined} functions, meaning it can freely decide how a given task should be divided. Then, undefined functions are detected, and the model is called again to define them after the fact. This bears some resemblance to Chain of Thought prompting \cite{li2023structuredcot}, but Liang \textit{et al.} find it can work even better in programming benchmarks such as HumanEval. LMP composition is not directly implemented they way it is used in CaP: LMPVC does not use different prompts to separately process parts of the voice command. Rather, the concept is a part of the inspiration for the Policy Bank. \cite{liang2023code}

\subsection{Policy Bank}

To extend the capabilities inherited from CaP, we propose the Policy Bank. Combining parts of the LMP composition and use of libraries demonstrated by Liang \textit{et al}, the Policy Bank is asolution for dynamically registering new Pythonic policies with LMPVC, and automatically incorporating them into the prompts sent to Code Generation. A programmer can simply provide a code file containing import statements, one or more function definitions, and a short common sense hint, separated by special comment blocks. These files are then registered with the system through a JSON configuration file, which the Policy Bank reads to determine which files should be included when building the prompt for Code Generation.

Additionally, the Policy Bank interacts with the Core's keyword system to implement a voice-based teaching method. When the user says a specific keyword (by default: "record policy"), Core enables an additional step to its main loop: Every time a result is received from Code Generation, it is also passed back to the Policy Bank through a special recording method. This code is then slightly reformatted in order to be saved as a part of a new policy. Once the user indicates the new policy is ready with another keyword (by default: "save policy"), the Core asks the user for a name and a hint, which are also passed to the Policy Bank. With this information, a new policy can be created, and incorporated into the system in exactly the same way as a handwritten one would be. These learned policies are also saved to disk in the same text format, allowing further editing by a programmer.

\section{Results}
The primary goals of our experiments are to (i) show how LMPVC as a whole performs with common natural language voice commands, (ii) demonstrate how the \textit{Policy Bank} can be used to apply CaP concepts to specialized industrial use cases, and (iii) evaluate how this combination performs in practical tasks. Experiments \ref{sec:direct}, \ref{sec:policies} and \ref{sec:assembly} are mostly from the original M.Sc. thesis \cite{lmpvcThesis}, with additional ones conducted for this paper.

\subsection{Experimental setup}

The setup used for practical experiments consists of a \textbf{Franka Emika Panda} 7-DOF robot with a hand-mounted \textbf{Intel RealSense D435} camera, \textbf{three PCs} running the various ROS components, and an \textbf{industrial pump} used for collaborative assembly tasks. PC 1 is dedicated to controlling the robot, and needs no special hardware. PC 2 performs object detection using an Nvidia GTX 1080Ti 11GB GPU, while PC 3 contains an Nvidia RTX 3080 10GB GPU powering both StarCoder2-15B, and Whisper Small. The object detection implementation used for testing is based on the one developed by Ojanen for a thesis project \cite{ojanen2024hrc}.

\subsection{Direct robot control}\label{sec:direct}

Starting simply, \textbf{LMPVC can respond to various simple directives}. This includes parsing common units of measurement, based on inferring from examples that the base unit of the coordinate system is a meter. Code generation is also \textbf{resilient to some mistakes}. In the case of \textit{"Two... twenty centimeters to the left"}, the user's hesitation is transcribed by Whisper as \textit{"twotwenty"}. Fortunately, this is resolved correctly when it comes to setting the coordinate in code.

\begin{lstlisting}[language=Python, backgroundcolor=\color{hyellow}, belowskip=0pt]
USER: Two... twenty centimeters to the left.
\end{lstlisting}
\begin{lstlisting}[language=Python, aboveskip=0pt]
LMPVC:
def twotwenty_centimeters_to_the_left(robot):
    waypoint = robot.get_pose()
    waypoint.position.y -= 0.2
    robot.add_waypoint(waypoint)
    robot.go()
\end{lstlisting}

Similarly, it \textbf{can resolve ambiguity and imprecision} by making reasonable assumptions, such as interpreting moving \textit{a little} as 5 centimeters. Assumptions can also be directed by including some of them in the examples.

\pagebreak

\begin{lstlisting}[language=Python, backgroundcolor=\color{hyellow}, belowskip=0pt]
USER: Move a little down.
\end{lstlisting}
\begin{lstlisting}[language=Python, aboveskip=0pt]
LMPVC:
def move_a_little_down(robot):
    waypoint = robot.get_pose()
    waypoint.position.z -= 0.05
    robot.add_waypoint(waypoint)
    robot.go()
\end{lstlisting}

As with CaP, \textbf{sensors can be integrated} through their own Python interfaces. Since our solution uses a simple object detection solution, our examples direct code generation to only pass names of objects, which are then mapped to classes and converted to poses. The high-level interface is defined by the prompt, simplifying integration of different sensors.

\begin{lstlisting}[language=Python, backgroundcolor=\color{hyellow}, belowskip=0pt]
USER: Can you see the big bolt?
\end{lstlisting}
\begin{lstlisting}[language=Python, aboveskip=0pt]
LMPVC:
def can_you_see_the_big_bolt(robot):
    (big_bolt_pose, big_bolt_found) = robot.find('big_bolt')
    if(not big_bolt_found):
        robot.say("Can't find the big bolt!")
    else:
        robot.say("Found the big bolt!")
\end{lstlisting}

The most common occurrence of hierarchical generation within LMPVC is in the context of \textbf{multi-stage commands}. Separate functions are often called for sub-tasks, and subsequently generated. The output processing of our solution always places the top-level function at the bottom to make generated code more readable. 

\begin{lstlisting}[backgroundcolor=\color{hyellow}, belowskip=0pt]
USER: Move a little down, and then draw a circle with radius 35 millimeters.
\end{lstlisting}
\begin{lstlisting}[language=Python, aboveskip=0pt]
LMPVC:
def draw_circle_of_radius_35_millimeters(robot):
    waypoints = 25
    for i in range(waypoints + 1)):
        waypoint = robot.get_pose()
        waypoint.position.x += (0.035 * math.cos((((2 * math.pi) * i ) / 25)))
        waypoint.position.y += (0.035 * math.sin((((2 * math.pi) * i ) / 25)))
        robot.add_waypoint(waypoint)
    robot.go()

def move_a_little_down(robot):
    waypoint = robot.get_pose()
    waypoint.position.z -= 0.05
    robot.add_waypoint()
    robot_go()

def move_a_little_down_and_then_draw_a_circle...
    _with_radius_35_millimeters(robot):
    move_a_little_down(robot)
    draw_a_circle_of_radius_35_millimeters(robot)
\end{lstlisting}

\subsection{Context and conversational modification}

By retaining earlier code as context, LMPVC \textbf{can react to or modify previous behaviour}. For example, context can influence how ambiguities are solved, making it possible to use commands like \textit{"Just a bit more"} immediately after the previous \textit{"Move a little down"}. Some more advanced modifications are also possible, such as directing the robot to regenerate an action with different parameters. This functionality is reminiscent of examples Vemprala \textit{et al.} \cite{vemprala2023chatgptroboticsdesignprinciples} use to discuss the potential of the conversational capabilities of ChatGPT. 

\begin{lstlisting}[language=Python, backgroundcolor=\color{hyellow}, belowskip=0pt]
USER: Draw a small circle.
\end{lstlisting}
\begin{lstlisting}[language=Python, aboveskip=0pt, belowskip=0pt]
LMPVC:
def draw_a_small_circle(robot):
    waypoints = 25
    for i in range(waypoints + 1)):
        waypoint = robot.get_pose()
        waypoint.position.x += (0.05 * math.cos((((2 * math.pi) * i ) / 25)))
        waypoint.position.y += (0.05 * math.sin((((2 * math.pi) * i ) / 25)))
        robot.add_waypoint(waypoint)
    robot.go()
\end{lstlisting}

\begin{lstlisting}[language=Python, backgroundcolor=\color{hyellow}, aboveskip=0pt, belowskip=0pt]
USER: Double the radius.
\end{lstlisting}
\begin{lstlisting}[language=Python, aboveskip=0pt]
LMPVC:
def double_the_radius(robot):
    waypoints = 25
    for i in range(waypoints + 1)):
        waypoint = robot.get_pose()
        waypoint.position.x += (0.1 * math.cos((((2 * math.pi) * i ) / 25)))
        waypoint.position.y += (0.1 * math.sin((((2 * math.pi) * i ) / 25)))
        robot.add_waypoint(waypoint)
    robot.go()
\end{lstlisting}

The StarCoder2 LLM core used by this version of LMPVC is unquestionably more limited, but it is interesting that even far smaller models not fine-tuned for conversation can still be used in this manner.

\subsection{Policy bank}\label{sec:policies}

\textbf{Policies may be programmed by hand} to explicitly define desired behaviour. This is convenient for users with programming experience, and allows adding functionality which the LLM struggles with or can't access. In the following example, a handover policy is created, using both object detection and movement commands through the \textit{High Level Controller}.

\begin{lstlisting}[language=Python]
import time
#BODY
def handover(robot):
    # Find predefined handover pose
    (handover_pose, success) = robot.find('handover')
    # Move to it
    robot.add_waypoint(handover_pose)
    robot.go()
    robot.say("In handover position, releasing in two seconds!")
    start = time.time()
    end = time.time()
    while(end - start < 2.0):
        time.sleep(0.1)
        end = time.time()
    robot.open_hand()
# HINT
# define function: give me the held item
def give_me_the_held_item(robot):
    handover(robot)
# end of function
\end{lstlisting}

Going forward, prompts to \textit{Code Generation} will now include the section below the \textit{\#HINT}-tag, allowing the function \textit{handover(robot)} to appear in generated code when appropriate. When they exist, import statements above the \textit{\#BODY}-tag are included as well.

\textbf{Policies can be taught} by recording generated policy output as the user guides the robot to complete a task with descriptive commands. Below is an automatically generated policy in response to the command \textit{"If you see a pipe and a cover, say all parts found, otherwise tell me what's missing."}, as saved to disk by the \textit{Policy Bank}.

\begin{lstlisting}[language=Python]
# Generated code
def if_you_see_a_pipe_and_a_cover_say_all_parts...
    _found_otherwise_tell_me_whats_missing(robot):
    (pipe_pose, pipe_found) = robot.find('pipe')
    if (not pipe_found):
        robot.say("Can't find the pipe!")
    else:
        (cover_pose, cover_found) = robot.find('cover')
        if (not cover_found):
            robot.say("Can't find the cover!")
        else:
            robot.say('All parts found!')
# Added by Policy Bank based on user input
def parts_check(robot):
    if_you_see_a_pipe_and_a_cover_say_all_parts...
    _found_otherwise_tell_me_whats_missing(robot):
# HINT
# define function: inspect all parts
def inspect_all_parts(robot):
    parts_check(robot)
# end of function
\end{lstlisting}

\textbf{Policies can also include multiple commands} and even \textbf{references to existing policies}. Here, a policy is constructed from three commands: \textit{"Find the assembly and move thirty centimeters above it."}, \textit{"Check parts."} and \textit{"Check bolts."}. The first command generates new behavior, while the other call existing policies, including a previous example.

\begin{figure*}[tp]
    \centering
    \includegraphics[width=\textwidth]{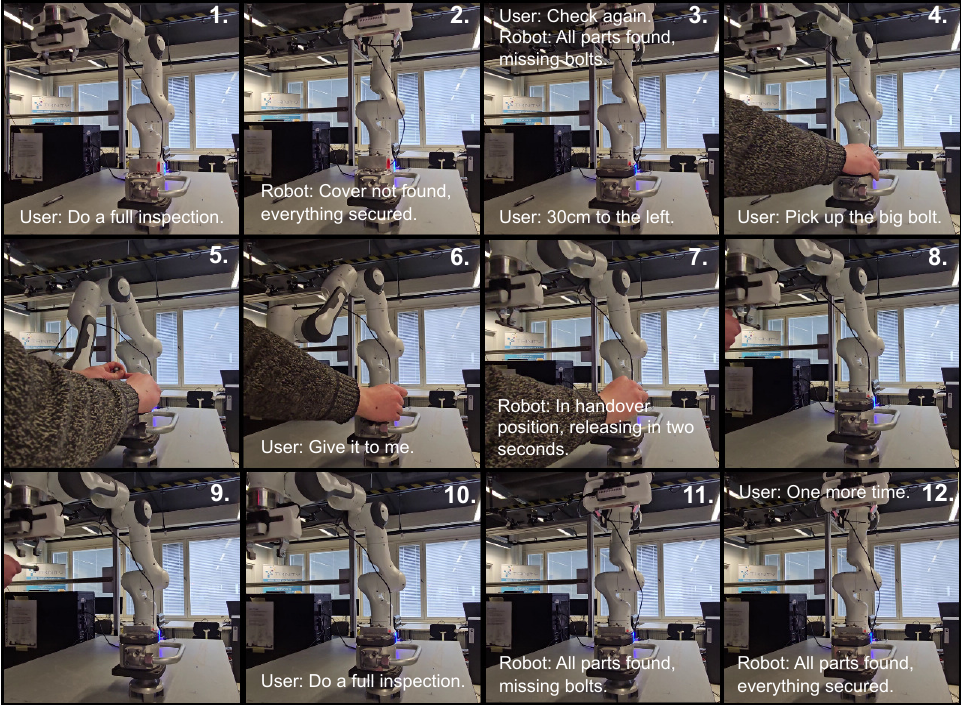}
    \caption{Experimental assembly of a pump, demonstrating practical human-robot collaboration aided by the Policy Bank. \cite{lmpvcThesis}}
    \label{fig:assembly}
\end{figure*}

\begin{lstlisting}[language=Python]
# Generated code based on 1st command
def find_the_assembly_and_move_thirty_...
    centimeters_above_it(robot):
    (assembly_pose, assembly_found) = robot.find('assembly')
    if (not assembly_found):
        robot.say("Can't find the assembly!")
    else:
        waypoint = robot.get_pose()
        waypoint.position.x = assembly_pose.position.x
        waypoint.position.y = assembly_pose.position.y
        waypoint.position.z = (assembly_pose.position.z + 0.3)
        robot.add_waypoint(waypoint)
        robot.go()       
# Generated code based on 2nd command
def check_parts(robot):
    parts_check(robot)   
# Generated code based on 3rd command
def check_bolts(robot):
    bolts_check(robot)
# Added by Policy Bank based on user input
def full_check(robot):
    find_the_assembly_and_move_thirty...
    _centimeters_above_it(robot)
    check_parts(robot)
    check_bolts(robot)
# HINT
# define function: do a full inspection
def do_a_full_inspection(robot):
    full_check(robot)
# end of function
\end{lstlisting}

Multiple commands and reusing existing policies combine to make the \textit{Policy Bank} a powerful tool for extending the capabilities of the LLM with robust and potentially very complex behaviour blocks. With the further addition of hand-written components and the ability to fix or modify the blocks later, the resulting architecture becomes highly customizable and less reliant on the capabilities of LLMs alone. If desired, the complete policies could even be triggered by some other, more robust voice control method, leaving the LLM only as a teaching tool and secondary control method.

\subsection{Pump assembly}\label{sec:assembly}

\begin{figure}[h!]
    \centering
    \includegraphics[width=\linewidth]{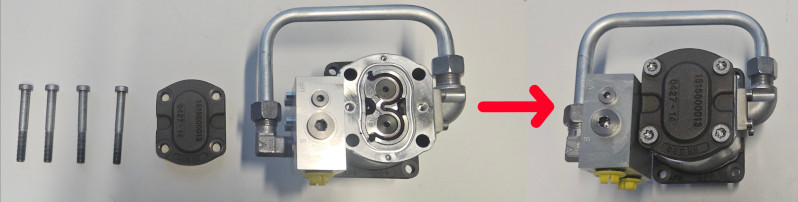}
    \caption{The goal of the collaborative task.}
    \label{fig:parts}
\end{figure}

To demonstrate how LMPVC and the Policy Bank can enable HRC in practice, we conduct an experimental assembly task: A human and a robot arm collaboratively assembling a pump. For brevity, the experiment focuses on the last step of this assembly, which consists of \textbf{installing the top cover} and \textbf{securing it with four screws}, as shown in Fig. \ref{fig:parts}. This task makes use of the policies created in the previous section, along with some more general commands. Screenshots of the experiment are shown in Fig. \ref{fig:assembly}, and a full video is provided on GitHub. 

First, the user asks the robot to \textit{"Do a full inspection"}, \textbf{causing code generation to call the generated policy} from the end of the previous section. Note that this policy also checks for the presence of a \textit{pipe}, which at this stage is already attached to the side of the pump. After the policy is executed, the robot responds with \textit{"Cover not found, everything secured."}, and the user promptly places the cover on top of the assembly. Here, the \textbf{context sensitivity} of LMPVC becomes useful, as the user can trigger the inspection policy once more by simply saying \textit{"Check again."}. After listening to the updated response \textit{All parts found, missing bolts."}, the user moves the robot out of the way with \textit{"30cm to the left."}. 

While working on the assembly with both hands, the user can then direct the robot to perform a parallel task, sending it to \textit{"Pick up the big bolt."}. This \textbf{activates the hand-written policy} \textit{pick}, which takes care of how the robot approaches the target and picks it up. Similarly, the following command \textit{"Give it to me."} calls the \textit{handover} policy shown in the earlier examples. Finally, the user calls the inspection policy two more times with and without context, making the corrections the robot requests until the assembly is complete.

This experiment demonstrates that the Policy Bank can allow LMPVC to adapt to a practical HRC situation where purely AI-based alternatives might struggle. The inspection functionality was achieved through voice-based teaching, demonstrating how users with limited programming experience can make adaptations of their own.

\subsection{Code generation latency}

An additional concern for the practical viability of LMPVC is the time it takes for the local LLM to generate code. If the use of an LLM leads to excessive command latency, user experience could be significantly impacted. To quantify the impact, we measure the latency of the code generation component for a set of 50 commands ranging from movement and manipulation to more general knowledge-based tasks. A handful examples of these commands are listed in table \ref{tab:commands}, the rest can be found on GitHub with the other experiment materials.

\begin{table}[h!]
    \centering
    \begin{tabular}{|l|p{7.2cm}|}
    \hline
    N   & Command \\ \hline
    11. & Rotate 33 degrees clockwise. \\ \hline
    14. & Draw an ellipse with radii five and four centimeters. \\ \hline
    18. & What's the cube root of three rounded to two decimal places? \\ \hline
    31. & Move a little to the left and then right, repeat three times. \\ \hline
    50. & What tool should I use to tighten bolts to correct torque? \\ \hline
    \end{tabular}
    \caption{Examples of the commands used for the latency and accuracy measurements}
    \label{tab:commands}
\end{table}

Latency is measured 10 times for each command in the set, using a script with a short wait time between runs to avoid congestion. The results will vary between hardware and software configurations, and should not be taken as absolute values. Rather, the intent is to present some representative samples on relatively modern but still accessible hardware. This experiment is performed on an Nvidia RTX 3080 10GB, running StarCoder2-15B in GGUF IQ4\_XS format \cite{mradermacherdolphincoder}. The model is loaded with Llama.cpp \cite{llamacpp}, using FlashAttention-2 \cite{dao2023flashattention2fasterattentionbetter} for optimal performance.

\begin{figure}[h!]
    \centering
    \includegraphics[width=\linewidth]{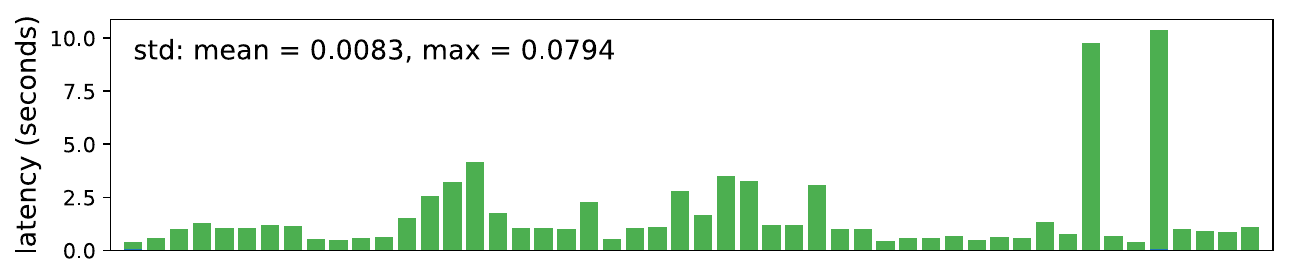}
    \caption{Mean command latency in the test set.}
    \label{fig:latency}
\end{figure}

The mean latency results for each command can be seen in Fig. \ref{fig:latency}, along with aggregate information of the standard deviations between runs for readability. It is evident that the results are consistent, with longer latencies resulting in the highest variability. More importantly, the mean latencies are quite low, below 1.5 seconds for majority of the test cases, and just over 4 seconds at maximum for successful commands. Latencies generally depend on the amount of code required to accomplish the task, corresponding to the \textit{tokens/s} performance of the model, and in this case two of the failures generated very long responses, resulting in higher latency. Even accounting for the minor additional latency from Whisper not included in this test, the successful results compare quite well to a simple, non-AI based voice control solution \cite{kaczmarek2020voicegestures}. This supports the viability of LLMs for real-time voice control, which we believe is promising for industrial applications.

\subsection{Command completion performance}

Due to the specialized focus of our work, LMPVC is not well suited to benchmarks such as RLBench, which primarily test performance in complex, open-ended manipulation tasks. In the absence of suitable standardized tools, we use the same 50 commands from the latency test to quantify reliability in a more appropriate scope. We choose test commands which are possible to complete with the tools the system has access to, with varying complexity. In order to make the results more meaningful, all commands were chosen and then tested at once. To collect results, we analyze the generated code for each command, calculate the success rate, and make note of relevant failures.

With this methodology, LMPVC successfully completes 39 out of 50 tasks, resulting in an overall success rate of \textbf{78\%}. The system generally does well in tasks involving moving the robot arm, calling policies and reacting to object detection, including making use of the received poses when required. It can also provide some advice, for example, the question \textit{"What tool should I use to tighten bolts to correct torque?"} is appropriately answered with \textit{"I would use a torque wrench to tighten bolts to the correct torque"}.

One notable failure involves the use of methods not present on the high-level controller. When asked to \textit{"press the red button"} the non-existent method \textit{robot.set\_digital\_output(0, True)} is called, indicating that the system failed to construct a pressing action from the known object detection and movement components. Similarly, something interesting happens in response to the command \textit{"Build a house"}. This is quite clearly an out-of-scope instruction, without any relevant policies to call. Often, the default answer of the model is to repeat the unknown task back at the user, but in this case, it attempts to build a house by picking and placing a variety of components, such as \textit{"bricks"}, \textit{"roof"} and \textit{"chimney"}. These examples highlight how the flexibility of AI can sometimes be an issue with unexpected commands.

In two further cases, the robot struggles to parse the phrase "back the other way", failing to move in the correct contextual direction. The LLM appears to often not understand the difference between "back" in the context of the previous answer, and "back" as in "backwards" in world space, although other contextual commands usually work quite well. The remaining six failures mostly involve incorrect information retrieval, and two cases of unexpected output on tasks similar to other, successful ones. These issues are more directly indicative of the limitations of a small LLM, although the 4-bit quantization of the model may also be a contributing factor. 

\section{Discussion}

Although LMPVC is at a disadvantage when relying purely on code generation, with the Policy Bank it can adapt to specialized industrial use cases not presently covered by Code as Policies \cite{liang2023code} or other similar code generation solutions \cite{singh2023progprompt}. Conversational behavior and contextual awareness reminiscent of the work of Vemprala \textit{et al.} \cite{vemprala2023chatgptroboticsdesignprinciples} can also be achieved with adjustments to prompting, despite the small model and lack of fine-tuning. Our solution additionally benefits from increased data security and low latency compared to works relying on larger cloud-based models, with results similar to certain non-AI alternatives \cite{kaczmarek2020voicegestures}.

We also identify several limitations, most commonly related to the capabilities of the local LLM, as well as the simplifications made to object detection and the architecture in general. Despite the advantages of augmenting live generation with existing policy, the system would still benefit from improved generation performance. While research on improved models continues to advance rapidly, we also suggest further investigating strategies for structured problem solving, such as LMP Composition \cite{liang2023code}, Chain of Thought and ReAct, as well as augmentation techniques such as Retrieval-Augmented Generation \cite{kang2024nadine}. The Policy Bank itself could additionally benefit from a robust solution for policies to send and receive data from each other and external processes, enabling more dynamic sequences of actions.

\section{Conclusion}\label{sec:conclusion}

In this paper, we proposed LMPVC, an open voice control architecture built for ROS 2, and Policy Bank, a framework for intelligently integrating voice-based teaching and high-level Python programming with LLM-based natural language robot control. In experiments involving individual commands, policy usage and an assembly task, this combination achieves promising performance using very limited hardware resources. While our solution does not compete with SotA methods in benchmarks measuring complex manipulation, it has potential advantages in industrial applications and other use cases requiring data security, low latency and adaptability to specialized tasks. 

\section*{Acknowledgements}

This project has received funding from the European Union's Horizon Europe research and innovation programme under grant agreement no. 101135708 (JARVIS). 

\bibliographystyle{IEEEtran}
\bibliography{IEEEabrv,refs}

\end{document}